# Human Disease Diagnosis Using a Fuzzy Expert System

Mir Anamul Hasan, Khaja Md. Sher-E-Alam and Ahsan Raja Chowdhury

**Abstract**—Human disease diagnosis is a complicated process and requires high level of expertise. Any attempt of developing a web-based expert system dealing with human disease diagnosis has to overcome various difficulties. This paper describes a project work aiming to develop a web-based fuzzy expert system for diagnosing human diseases. Now a days fuzzy systems are being used successfully in an increasing number of application areas; they use linguistic rules to describe systems. This research project focuses on the research and development of a web-based clinical tool designed to improve the quality of the exchange of health information between health care professionals and patients. Practitioners can also use this web-based tool to corroborate diagnosis. The proposed system is experimented on various scenarios in order to evaluate it's performance. In all the cases, proposed system exhibits satisfactory results.

—————— ◆ ——————

## 1 INTRODUCTION

ONLINE diagnosis is becoming popular day by day. In today's world people are so busy, that they hardly have enough time to visit a doctor. So they can use the service of this online diagnosis system residing home or office and have an idea about the disease. After that they can consult the specialist doctor if it is necessary or serious. This research presents a novel method for online diagnosis. In this method, based on the selection of the problem area/ problem, the expert system gives some symptom from which the user needs to select symptoms. Based on the selection of symptoms, the user is again asked some questions. According to the answer selection, the fuzzy expert system diagnosis diseases based on it's knowledge, add catalyst factor (if any), do ranking and gives the result in fuzzy form. As fuzzy expert system deals with uncertainty and vague terms, it is generously accepted in different sphere of life [1].

Rest of the paper is organized as follows- Section 2 highlights the background study. Proposed System is presented in Section 3 and evaluation of the proposed system is shown in Section 4. Chapter 5 concludes the paper.

## 2 BACKGROUND STUDY

Knowledge is a theoretical or practical understanding of a subject or a domain. In other words, Knowledge is the sum of what is currently known [2]. Diagnosis system is a system which can diagnose diseases through checking out the symptoms. A knowledge based online diagnosis system is developed for diagnosis of diseases based on the knowledge given by doctors in the system.

A computer Program Capable of performing at a human-expert level in a narrow problem domain area is called an expert system [2]. Management of uncertainty is an intrinsically important issue in the design of expert systems because much of the information in the knowledge base of a typical expert system is imprecise, incomplete or not totally reliable [3].

An expert system that uses fuzzy logic instead of Boolean logic is known as Fuzzy expert system.

A fuzzy expert system is a collection of fuzzy rules and membership functions that are used to reason about data. Using fuzzy expert system expert knowledge can be represented that use vague and ambiguous terms in computer.

Fuzzy logic is a set of mathematical principles for knowledge representation based on degrees of membership rather than the crisp membership of classical binary logic. Unlike two-valued Boolean logic, fuzzy logic is multi valued. Fuzzy logic is a logic that describes fuzziness. As fuzzy logic attempts to model human's sense of words, decision making and common sense, it is leading to more human intelligent machines [4].

Fuzzy logic was introduced in the 1930 by Jan Lukasiewicz; a Polish Philosopher (extended the truth values between 0 to 1). Later, 1937 Max Black defines first sample fuzzy set. In 1965, Lotfi Zadeh rediscovered fuzziness, identified and explored it [2].

The National Research Council, Canada has developed an online diagnosis system. Initially, they used acute leukaemia disease data to illustrate the efficacy of the diagnostic tool. More recently, they developed a new fuzzy classification method called PROAFTN [5] to assist medical diagnosis. Results obtained using the PROAFTN method on acute leukaemia diagnosis are promising. The challenge herein is to provide secure, remote access, to

————————
- *Mir Anamul Hasan is with the Dept. of Computer Science & Engineering (CSE), University of Dhaka, Dhaka, Bangladesh.*
- *Khaja Md. Sher-E-Alam is with the Dept. of CSE., University of Dhaka, Dhaka, Bangladesh.*
- *Ahsan Raja Chowdhury is with the Dept. of CSE, University of Dhaka, Dhaka, Bangladesh.*

JOURNAL OF COMPUTING, VOLUME 2, ISSUE 6, JUNE 2010, ISSN 2151-9617
HTTPS://SITES.GOOGLE.COM/SITE/JOURNALOFCOMPUTING/
WWW.JOURNALOFCOMPUTING.ORG
67decision support tools and a standard framework for the exchange of health information, over inexpensive Internet communication pathways using web-based technologies. Informantion related to this can be found from the website of National Research Council Canada [5].

YourDiagnosis [6] is developed by an expert panel of very experienced doctors in Australia. They are part of a large medical and hospital group called Macquarie Health Corporation Ltd. Macquarie Health Corporation Ltd. also owns Macquarie Hospital Services and Macquarie Medical Systems. Macquarie Hospital Services has owned and operated private hospitals since 1976.

Easy Diagnosis [7] is a online diagnosis system. This expert system software provides a list and clinical description of the most likely conditions based on an analysis of patient's particular symptoms. EasyDiagnosis focuses on the most common medical complaints that account for the majority of physician visits and hospitalizations.

Wrong Diagnosis [8] which is part of Health Grades network and WebMD [9] whose slogan is "Better Information, Better Health" are another two efficient online diagnosis tools.

There are many online diagnosis tool in the web, but proposed system is the most user friendly (easy to use), precise, at the same time accurate. In the web, some systems are so boring -needs to give lots of information, give answers to lots of question which is quite monotonous. In most of the other online diagnosis websites, the sequence of questions is fixed.

But in the proposed system one needs to give minimal information, questions will be asked after filtering at every step based on your responses and at every step the problem domain will become narrower to reach a perfect diagnosis.

## 3 PROPOSED FUZZY EXPERT SYSTEM FOR HUMAN DISEASE DIAGNOSIS

Different diagnosis strategies have been studied. Different e-health websites have been browsed. Papers and articles on e-health management have been read and different diagnosis tools have been used. Noted down important information like which information they are collecting from the user/ patients and which approach they are following in diagnosis.

Then information (symptom, signs etc.) of different diseases have been collected for particular category as sample [8, 9, 10]. Made a comparative analysis to identify which symptoms are major symptoms for particular diseases i.e. that symptom must be present, their acuity level etc.

Then met with few doctors with different level of expertise in their field. Asked them about those diseases and their symptoms, which factors they consider while they diagnosis diseases, what approach they follow.

After studying and making a comparative analysis of what is gotten by consulting with different doctors about which factors they consider while they diagnosis diseases, what approach they follow, it is found that they are almost same. Then a uniform structure is made and a mathematical equivalence (equation (3.1)) is formed, which will be used to diagnosis by the fuzzy expert system.

After that implementation of the expert system is done and tested it thoroughly to check whether the desired result is coming or not. After necessary corrections, when it seems ok with respect to the desired goal, then consulted with the doctors again and explained about the software implementation and showed it to them and they acknowledged it, gave some suggestions(to add catalyst factor which is some sort of personal information/ patient's history information), according to which the system is revised.

Probability of diseases is measured using equation (3.1) whose value depends on the user feedback while diagnosing.

Probability$_{disease}$ =

$$\frac{(\Sigma S_i * W_i - \Sigma (S_{major})_j (W_{max})_j) - Min_{th}}{Max_{th} - Min_{th}} * 100\% + \Sigma C_f$$

............ (3.1)

where-

i,j= counter of different symptoms
$S_i$ = is ith symptom selected or not
$W_i$ = weight of ith symptom
$(S_{major})_j$ = jth major symptom which is not selected by the user
$(W_{major})_j$ = weight of jth major symptom
$Max_{th}$ = maximum threshold of disease
$Min_{th}$ = minimum threshold of disease
$\Sigma S_i W_i$ = total weight of a disease based on the selected symptom
$\Sigma (S_{major})_j (W_{major})_j$ = total weight of unselected major symptoms of a disease
$C_f$ = catalyst factor based on patient's history/personal information

The proposed expert system assist user to diagnosis diseases, he/she might have, in a fuzzy way. Based on the selection of the problem area/problem, the expert system gives some symptom from which the user needs to select symptoms, based on the selection of symptoms the user is asked some questions, according to the answer selection the fuzzy expert system diagnosis diseases based on it's knowledge, add catalyst factor (if any) and do ranking and gives result in fuzzy form.

Fig. 1. represents the flow diagram of the proposed system.

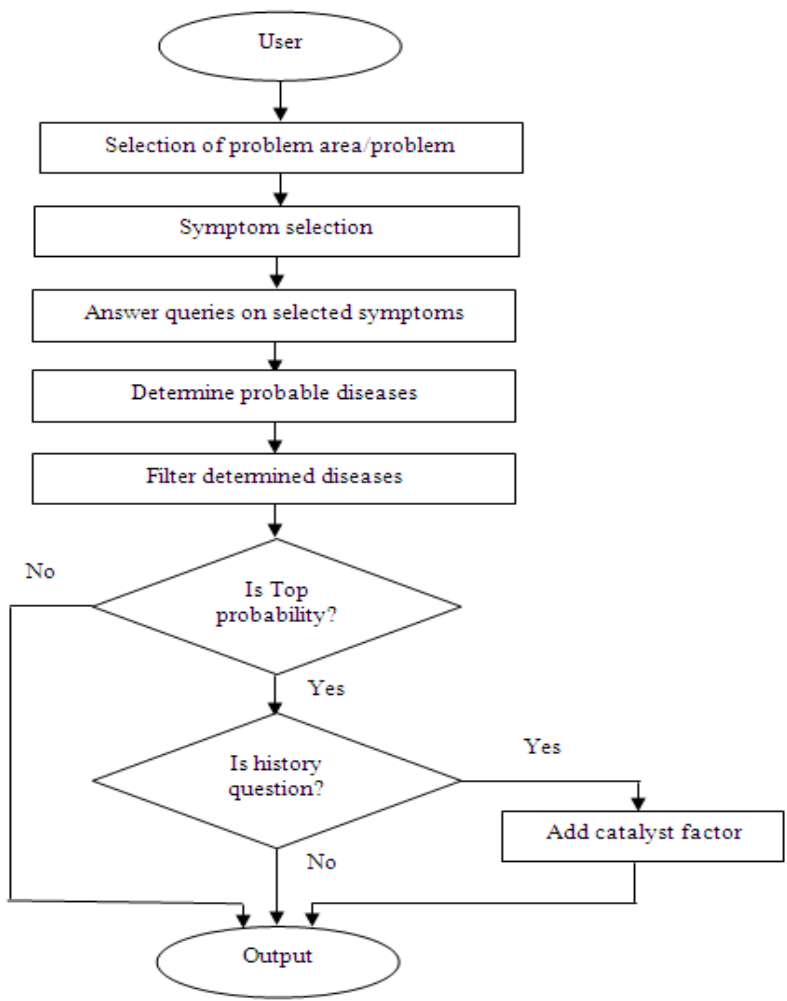

Fig. 1. Flow Diagram of the Proposed System.

## 4 EVALUATION OF THE PROPOSED SYSTEM

A computer Program Capable of performing at a human-expert level in a narrow problem domain area is called an expert system [1]. Basic characteristics of expert systems are [1]:

- Use knowledge rather than data
- Knowledge is encoded & maintained as an entity separate from the control program
- Capable of explaining how a particular conclusion is reached and why requested information is needed
- Use symbolic representation and perform inference through symbolic computation

Now if the characteristics of the proposed system are observed, it is seen that, it performs in a narrow problem domain area (only diagnosis diseases) at a human expert level (like doctors). It uses knowledge to make diagnosis decision; the knowledge is given by some expert doctors. It can also explain how a particular decision made, that is why the proposed system is an expert system. As it is using knowledge base in making decision, so it is also knowledge based expert system. The proposed knowledge base online diagnosis system also corresponds to fuzzy expert system as it simulates the fuzzy inference-process which is depicted in Fig. 2.

Let the user select chest as his/her problem area. Assume there are 3 diseases under category chest-Pneumonia, Bronchitis and Asthma. Then all the symptoms under chest category will be shown to the user by the expert system which are given in column one of Table 5.1. Then the user will select symptoms applicable to him/ her. After that the user will need to answer some question given by the expert system on selected symptoms. Based on the given answer (presented inside parenthesis of column one of Table 1) and answer of history question (if any- suppose in this case, the user need to answer two history question, in this particular case the catalyst factor is 4 and they belongs to Asthma) which is termed as catalyst factor, the expert system computes the probabilities of probable diseases and filter them before showing. The total diagnosis scenario is depicted in Table 1.

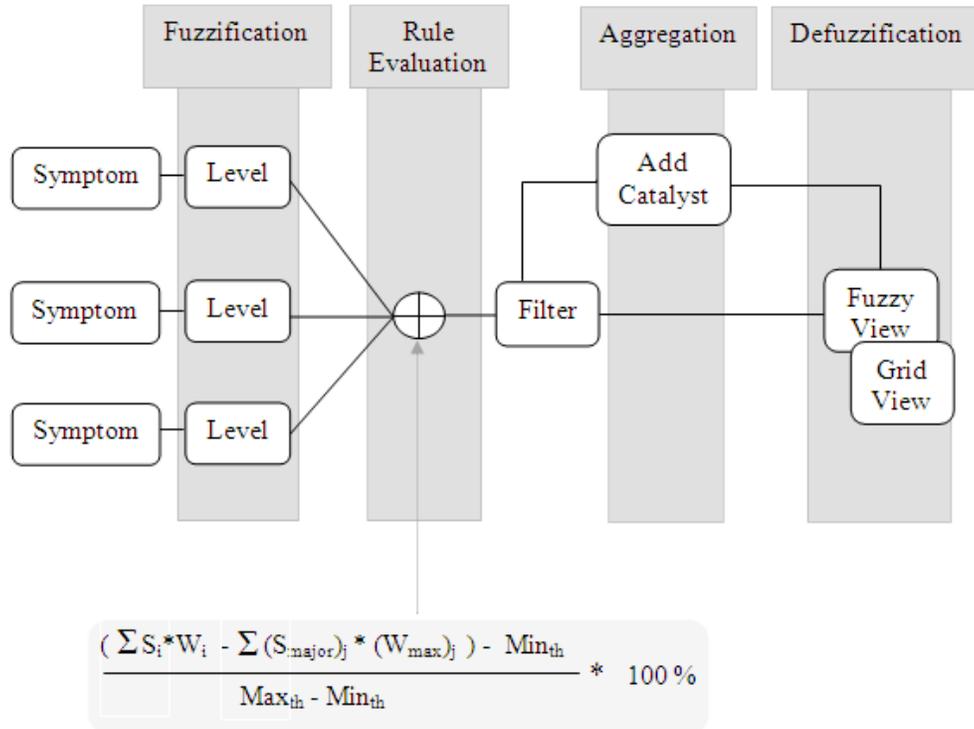

Fig. 2. Fuzziness of the proposed expert system.

TABLE 1
Units for Magentic Properties

| Symptom & their level | Pneumonia | Bronchitis | Asthma |
|---|---|---|---|
| Cough (non-productive) | 0.5 | 0.5 | 0.9 |
| Fever (low) | 0.1 | 0.7 | 0.0 |
| Chest Pain (always) | 0.3 | 0.5 | 0.2 |
| Wheezing Noise (while breathe in) | 0.0 | 0.9 | 0.9 |
| Hoarseness (didn't select) | 0.0 | 0.0 | 0.0 |
| Vomit (Never) | 0.0 | 0.0 | 0.0 |
| Short Breathe (yes) | 0.0 | 0.0 | 0.6 |

In case of Asthma: $\sum S_i W_i = 2.6$, $\sum (S_{major})_j (W_{major})_j = 0$, $Max_{th}=3.3$, $Min_{th}=0.2$ and $C_f = 4$
According to the Equation 3.1,

$$\text{Probability}_{Asthma} = \frac{2.6 - 0 - 0.2}{3.3 - 0.2} * 100\% + 4$$

$$= 81.419\%$$

In case of Bronchitis: $\sum S_i W_i = 2.6$, $\sum (S_{major})_j (W_{major})_j = 0$, $Max_{th}=4.5$, $Min_{th}=0.8$ and $C_f = 0$
According to the Equation 3.1,

$$\text{Prob.}_{Bronchitis} = \frac{2.6 - 0 - 0.8}{4.5 - 0.2} * 100\% + 0$$

$$= 48.648\%$$

In case of Asthma: $\sum S_i W_i = 0.9$, $\sum (S_{major})_j (W_{major})_j = 0$, $Max_{th}=3.6$, $Min_{th}=0.4$ and $C_f = 0$
According to the Equation 3.1,

$$\text{Prob.}_{Pneumonia} = \frac{0.9 - 0 - 0.4}{3.6 - 0.4} * 100\% + 0$$

$$= 15.625\%$$

**Rule:**

IF        Probability$_{disease}$(temp) is Max AND
          Catalyst Factors($C_f$) are Present
THEN      Probability$_{disease}$ = Probability$_{disease}$(temp) + $\Sigma C_f$
ELSE      Probability$_{disease}$ = Probability$_{disease}$(temp)

The following chart (Fig. 3) gives a quick overview of the confidence level of the proposed system with real time diagnosis. The bar on right for every disease indicates the confidence level of our system without having observation on any pathological test results. And the bar on left shows the full confidence of diagnosis in real time with observing all necessary pathological tests. In our system the confidence level decreases with the increase of number of pathological tests needed.

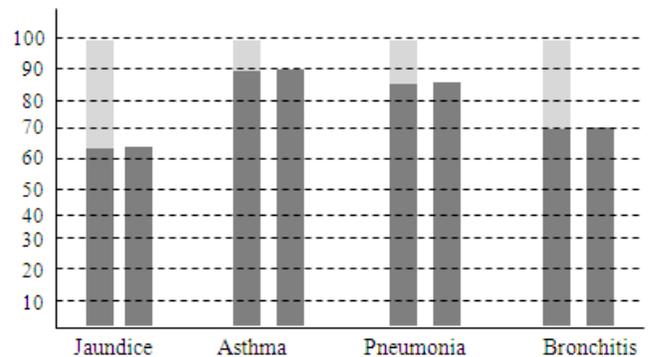

Fig. 3. Confidence level of the proposed system.

## 4 CONCLUSION

E-applications play a dominant role in the realm of overwhelming Internet technology. Knowledge based applications are the features of latest online technology. Proposed knowledge based online diagnosis system can play a vital role for the users of the system. The system facilitates the users to determine his/ her probable diseases very quickly with the aid of a knowledge based expert system. As the knowledge base is created based on the feedback from expert and specialist doctors, users can also rely on it.

### ACKNOWLEDGMENT

The authors would like to express their sincere thanks to Prof. Dr. Md. Hasanuzzaman, Dr. Abdur Rahman, Dr. Hasibul Hasan Parag and Dr. Salauddin Ahmed for their collaboration.

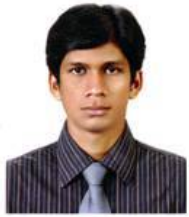
**Mir Anamul Hasan** is a graduate student in Department of Computer Science & Engineering, University of Dhaka, Bangladesh. His research interests include Artificial Intelligence, Fuzzy Logic, Reversible Logic, Bioinformatics etc.

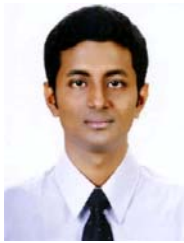
**Khaja Md. Sher-E-Alam** is a graduate student in Department of Computer Science & Engineering, University of Dhaka, Bangladesh. His research interests include Artificial Intelligence, Fuzzy Logic, Reversible Logic, Bioinformatics etc.

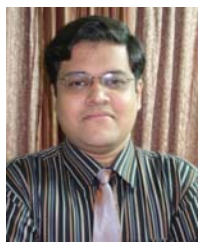
**Ahsan Raja Chowdhury** is working as an Assistant Professor in Department of Computer Science & Engineering, University of Dhaka, Bangladesh. He has received his BSc (Hons) and MS from the same institution. He has already published few International Journal and Conference papers and participated some International Conferences. His research interests include Artificial Intelligence, Fuzzy Logic, Reversible Logic, Bioinformatics, etc.